\documentclass[sigconf]{acmart}
\AtBeginDocument{%
  }

\setcopyright{acmlicensed}
\copyrightyear{2018}
\acmYear{2018}
\acmDOI{XXXXXXX.XXXXXXX}
\acmConference[Conference acronym 'XX]{Make sure to enter the correct
  conference title from your rights confirmation email}{June 03--05,
  2018}{Woodstock, NY}
\acmISBN{978-1-4503-XXXX-X/2018/06}

\usepackage{comment}
\usepackage{url}
\usepackage{booktabs}
\usepackage{subcaption}
\usepackage{multirow} 
\usepackage[most]{tcolorbox}
\usepackage{footmisc}

\newtcolorbox{insightbox}[1]{
    colback=orange!10,      
    colframe=orange!70, 
    colbacktitle=orange!70, 
    title={#1},             
    fonttitle=\sffamily\bfseries\large,
    fontupper=\sffamily,    
    arc=15pt,               
    outer arc=15pt,
    left=15pt,              
    right=15pt,
    top=10pt,
    bottom=10pt,
    boxrule=1.5pt,          
    titlerule=0pt,          
    toptitle=5pt,
    bottomtitle=2pt,
    enhanced,               
}
\begin{document}


\title{Emotion-Aware Clickbait Attack in Social Media}

\author{Syed Mhamudul Hasan}
\affiliation{%
  \institution{Southern Illinois University}
  \city{Carbondale}
  \state{IL}
  \postcode{62901}
  \country{USA}}
\email{syedmhamudul.hasan@siu.edu}

\author{Mohd. Farhan Israk Soumik}
\affiliation{%
  \institution{Southern Illinois University}
  \city{Carbondale}
  \state{IL}
  \postcode{62901}
  \country{USA}}
\email{mohdfarhanisrak.soumik@siu.edu}

\author{Abdur R. Shahid}
\affiliation{%
  \institution{Southern Illinois University}
  \city{Carbondale}
  \state{IL}
  \postcode{62901}
  \country{USA}}
\email{shahid@cs.siu.edu}

\renewcommand{\shortauthors}{Hasan et al.}



\begin{abstract}
Clickbait is characterized by disproportionately high emotional intensity relative to informational content, often reinforced by specific structural patterns. However, current research considers clickbait as a static textual phenomenon characterized by linguistic patterns and structural cues. Additionally, existing detection systems primarily rely on surface-level features of clickbait. This paper introduces an emotion-aware clickbait generation attack, where stylistic transformations are used to optimize emotional impact. We propose an emotion-aware framework based on the Valence–Arousal–Dominance (VAD) space to model the emotional dynamics underlying clickbait generation for optimal user engagement. To simulate realistic attack scenarios, we align clickbait headlines with semantically similar social media posts using Sentence-BERT and generate multiple stylistic rewrites via Large Language Models (LLMs). Building on this, we define a Curiosity Gap (CG) function that computes clickbait's headline variation to the current post to quantify how emotional activation will contribute to user curiosity and evade the existing system found on social media. Experimental results demonstrate that emotion-aware stylization significantly degrades the performance of state-of-the-art classifiers, leading to misclassification rates of up to 2.58\% to 30.63\% on the base system.
\end{abstract}

%

\received{20 February 2007}
\received[revised]{12 March 2009}
\received[accepted]{5 June 2009}
\maketitle

\section{Introduction}







Clickbait has become a dominant mechanism for maximizing engagement in digital media by exploiting user attention through emotionally charged and information-incomplete content~\cite{jung2022click, immorlica2024clickbait}. Unlike traditional misinformation, which presents factually false content, clickbaiting operates through a more subtle mechanism: using eye-catching but often misleading headlines or descriptions to encourage users to click on content, thereby generating ad revenue and increasing click-through rates~\cite{shrestha2024first}. This content tends to focus on sensational, personalized, and entertaining topics, while non-clickbait content is more aligned with public affairs and informative news. Clickbait shares similarities with misleading content but differs in that it manipulates attention primarily through exaggeration and curiosity rather than factual falsification~\cite{shin2025emotion}.










In emerging agentic media, such as multi-agent social platforms named Moltbook, clickbait evolves from a static content property into a dynamic strategy, where agents adaptively generate emotionally optimized content to influence interaction and maximize engagement. In Moltbook, an agent can post multiple automated clickbait with the contextual understanding with the post. An agent deliberately generates emotionally manipulative, incomplete, or exaggerated content to influence other agents’ behavior and maximize engagement. This introduces risks such as coordinated emotional manipulation, attention hijacking, and large-scale amplification of misleading content.

Social media platforms have implemented a range of strategies to address clickbait and misinformation. For example, flagging tweets as originating from bot accounts~\cite{abokhodair2015dissecting} or as containing misinformation has been shown to reduce users' positive attitudes toward such content, although this effect is weaker among individuals who use social media frequently~\cite{ecker2022psychological}.  Similarly, LLMs have demonstrated substantial capabilities in detecting clickbait contents~\cite{muqadas2025deep}. Through reinforcement learning with human feedback, LLMs can increase the effectiveness of its detection system~\cite{du2024reinforcement}.

Despite significant progress in clickbait detection and mitigation by LLM, several important challenges remain~\cite{wang2025clickbait, ai2026paradigm}. 
Existing approaches to clickbait detection primarily focus on surface-level linguistic features, such as lexical patterns and headline structures. However, these methods overlook a critical dimension: the role of emotional manipulation in driving engagement. Prior studies show that clickbait relies heavily on curiosity gaps, exaggeration, and emotional triggers, yet these factors are rarely modeled explicitly. In this study, we propose an emotion-aware formulation of clickbait, where clickbait is modeled not only as a textual artifact but as an emotion-driven behavioral strategy that shapes interaction dynamics over time. We use a strategic form of digital content that leverages curiosity gaps, emotional stimulation, exaggeration, and incomplete information to maximize user engagement (e.g., clicks, views, interactions), often at the cost of content quality or accuracy~\cite{jacobo2024clickbait}. The headline plays a pivotal role in clickbait strategy, as it is designed to capture attention and provoke curiosity, encouraging readers to engage with the full article or share it on social media~\cite{bazaco2019clickbait, wang2025decoding}.

This work addresses the following research questions for this proposed attack based on stylistic transfer of clickbait:

\begin{itemize}
    \item \textbf{RQ1:} How to build emotion aware clickbaiting?
    \item \textbf{RQ2:} How many different ways can we apply the emotion to create clickbait through the VAD space?
    
\end{itemize}

\begin{figure}[t!]
    \centering
    \includegraphics[width=\columnwidth]{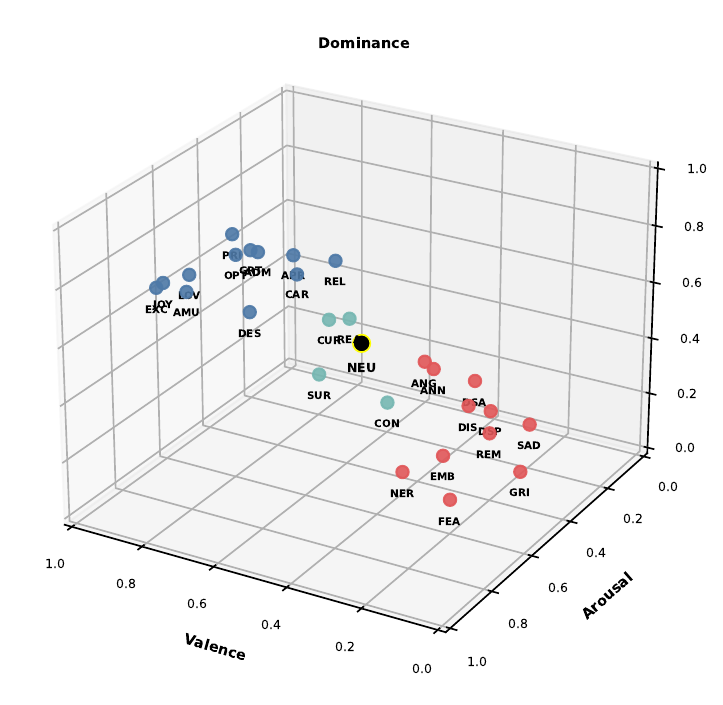}
    \caption{Emotion mapping in Valence-Arousal-Dominance (VAD) Framework}
    \label{fig:vad}
\end{figure}

\section{Background}

\subsection{Stylistic Transformation}

Stylistic transformation refers to modifying the expression of a message while preserving its original meaning~\cite{reif2022recipe}. LLMs can perform stylistic transformation through prompt engineering by rewriting text into different tones while preserving semantic meaning. %

\subsection{Social Amplification Attack}

Social amplification attack is a strategy in which an adversary generates emotionally optimized clickbait content to amplify user engagement and influence interaction dynamics in social networks by stylistic transformation. This attack can be devised by modern social media platforms, where both post titles and content are publicly accessible. An attacker can leverage this accessible information to analyze the underlying semantic and emotional structure of a post and strategically craft a response that maximizes its impact. Specifically, the attacker increases the curiosity gap by exploiting both the title and the post content, selectively emphasizing emotionally activating and informationally incomplete aspects. To ensure contextual relevance, the attacker identifies appropriate clickbait content through semantic similarity, aligning the generated text with the target post. This alignment is performed in the embedding space, enabling the attacker to inject semantically coherent yet emotionally amplified content into the comment section of the original post. As a result, the injected content appears contextually relevant while simultaneously increasing emotional activation and reducing informational closure, thereby amplifying user attention and interaction. This behavior transforms clickbait from a passive content property into an active amplification mechanism within social networks.

\subsection{Related Work}

The proliferation of clickbait on digital platforms has become a major issue that disrupts accurate reporting and erodes public confidence in online content~\cite{munger2020all, thinnakkakath2025social}. Social media platforms, with their emphasis on content virality and engagement metrics, have created an ideal environment for clickbaiting. The economic incentives underlying clickbait creation are particularly important to understand: online publishers exploit attractive news headlines as a way to create advertisements, with spammers capitalizing on the attention-grabbing nature of sensationalized content~\cite{ahmed2025impact}. This practice, while profitable, compromises the credibility of digital media and undermines informed public discourse ~\cite{devi2025truth}.

To detect and understand the nature of clickbait, Mackey et al.~\cite{mackey2025emotion} proposed a novel framework by focusing on emotional manipulation rather than just linguistic patterns using the Valence Arousal Dominance (VAD) emotion space, where effective clickbait typically combines high emotional intensity (arousal), strong positive or negative sentiment (valence), and low informational clarity (dominance) to create a curiosity gap. This curiosity gap explains why users feel compelled to click. The study further integrates semantic similarity techniques to ensure that generated clickbait remains relevant to the original content while maximizing emotional impact. Rony et al.~\cite{rony2017diving} showed that the clickbait posts generally generate higher user engagement, such as more reactions, shares, and comments, although they may reduce trust and credibility over time. To detect such text, Naeem et al.~\cite{naeem2020deep} proposed a deep learning framework for clickbait detection on social media by combining linguistic feature analysis with a modified LSTM model, showing that clickbait can be effectively identified through natural language cues such as curiosity gaps, pronoun usage, and sensational phrasing. Liu et al.~\cite{liu2023emotion} also proposed an emotion-aware framework for fake news detection that integrates textual content, emotional signals, and propagation structure within a unified graph-based model. Despite its strong performance, the paper primarily focuses on detection of fake news after it has been generated and propagated, rather than modeling how emotionally manipulative content is actively created or injected into social networks. Additionally, the framework assumes static propagation structures and does not fully address dynamic, real-time interaction scenarios or semantic alignment between injected content and existing posts. These limitations suggest a gap in understanding emotion-driven content generation as an active amplification mechanism, which this work addresses by framing clickbait as a social amplification attack using semantic similarity and curiosity gap modeling. Further, Kaushal et al.~\cite{kaushal2021clickbait} showed that clickbait headlines attract clicks but reduce news credibility and are influenced by user traits like age and curiosity. However, there is a gap of modeling clickbait as an emotion-driven, context-aware, and adversarial generation process rather than just studying its effects. 

Existing work~\cite{scott2021you} also explained clickbait as a linguistic mechanism that creates curiosity gaps, but it fails to model clickbait as an emotion-driven, context-aware, and adversarial generation process within social interactions in digital media.

\begin{insightbox}{Contribution.}

In this paper, we aim to understand clickbait in modern social networks by introducing a novel formulation of clickbait. We define a social amplification attack as a strategy in which an adversary generates emotionally optimized clickbait content to amplify user engagement and influence interaction dynamics in social networks. This framework defines a Curiosity Gap (CG) function to formally measure how emotional activation, informational incompleteness, and sentiment jointly drive user curiosity and interaction. Finally, we frame clickbait generation as a social amplification attack, where an adversary generates emotionally optimized and semantically aligned content to influence interaction dynamics. To support this formulation, we incorporate a semantic similarity–based alignment strategy that pairs generated clickbait with relevant social media posts, ensuring contextual realism in adversarial injection scenarios.
\end{insightbox}

\begin{figure*}[ht!]
    \centering
    \includegraphics[width=\textwidth]{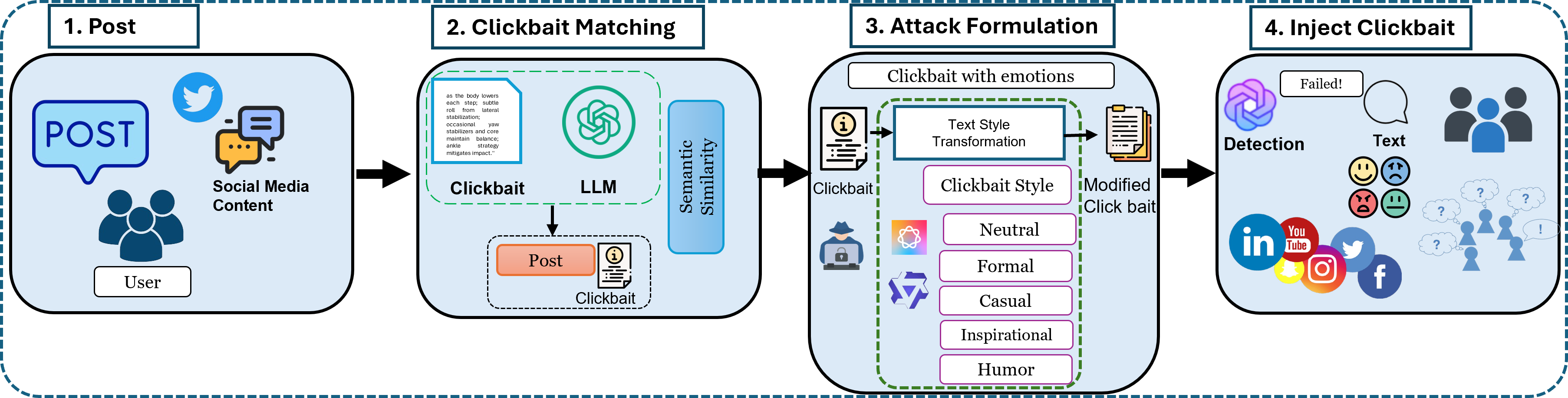}
    \caption{Overview of the emotion-aware clickbait generation and social amplification attack pipeline 
    }
    \label{fig:schema}
\end{figure*}

\section{Attack Model}

\textbf{\quad Assumptions.} We assume that the attacker operates under a black-box setting with no access to internal platform ranking or moderation mechanisms. The attack relies solely on publicly available textual information such as posts and comments. We also consider a social media environment where user-generated content, including post titles and associated text, is publicly accessible. These posts serve as the primary contextual signals for user interaction and engagement. We consider an adversary whose objective is to maximize user engagement by generating emotionally optimized clickbait content. The adversary does not modify the original post but operates by injecting clickbait content (e.g., comments) into the discussion thread or publish clickbait content on similar discussing community.

\textbf{Adversary Capabilities.} The attacker has access to publicly available post titles and content, which are used to infer semantic and emotional context. Leveraging this information, the attacker can analyze the post to identify emotionally salient and informationally incomplete aspects. Then, he generates clickbait-style content that increases the curiosity gap, and compute semantic similarity between candidate clickbait texts and the target post to ensure contextual alignment.

\textbf{Attack Strategy.} Given a target post, the attacker generates a clickbait or select suitable text such that it is similar to post. This is achieved by maximizing semantic similarity in embedding space. thereafter, the attacker applies the curiosity gap in that text and inject the generated content into the comment section of the post.

\textbf{Adversary Goal.} The goal of the attacker is to maximize attention and interaction e.g., clicks, replies, or engagement by increasing the curiosity gap and emotional activation of users while maintaining sufficient semantic consistency to avoid detection. The attacker also selectively aligns clickbait with semantically relevant posts and injects it into ongoing discussions, increasing the curiosity gap and emotional activation of users, thereby maximizing attention and interaction.


\section{Methodology}

\subsection{Curiosity Gap }

Clickbait works by leveraging a psychological concept known as the Curiosity Gap~\cite{dan2020clickbaits}. The clickbait headlines attract users by creating an information gap through specific linguistic features such as pronouns and superlatives, which trigger curiosity and exploit human cognitive tendencies to seek relevant information, as modeled using relevance theory.  

\section{Analysis in VAD Space}
To analyze the emotional structure of clickbait, we define the \textit{Curiosity Gap} ($CG$) in the Valence Arousal Dominance (VAD) space. Let a headline $h$ be represented by a VAD vector where $V_h$ denotes valence, $A_h$ denotes arousal, and $D_h$ denotes dominance, then we formalize
\[
v_h = (V_h, A_h, D_h), \qquad V_h, A_h, D_h \in [0,1],
\]

The proposed score represents a continuous measure of a headline’s curiosity-inducing potential where:
\begin{itemize}
    \item high $A_h$ increases attentional activation,
    \item low $D_h$ increases ambiguity or lack of closure,
    \item high $V_h$ reinforces engagement through positive emotional framing.
\end{itemize}

We treat the post emotion as the affective ground truth of the underlying content, and evaluate whether stylized headline rewrites amplify curiosity beyond the post’s emotional baseline while preserving semantic consistency.

Let $p$ denote the original post and let $c^{(s)}$ denote a clickbait-stylized comment under style $s$. We map each text into the VAD space using
\[
\phi(x) = (V_x, A_x, D_x),
\]
where $V_x$, $A_x$, and $D_x$ denote valence, arousal, and dominance, respectively.

The valence term $V_h$ is incorporated as an additive component to reflect the tendency of positively framed content to encourage approach-oriented engagement. Under this interpretation, a headline with higher valence can amplify the likelihood of user interaction, even when the curiosity signal is primarily driven by arousal and reduced dominance. Thus, higher values of $CG(h)$ indicate stronger curiosity-inducing potential. In practice, clickbait-oriented rewrites are expected to yield higher $CG$ values than neutral or fully informative headlines. Conversely, de-clickbaited or informational rewrites should reduce $CG$ by lowering arousal, increasing dominance, or both. To measure the emotional shift between the post and the stylized clickbait comment, we define the VAD-space. In addition, to quantify the increase in curiosity induced by the stylized comment, we define the curiosity gap score for a text $x$ as
\[
\boxed{
CG(x) = A_x(1-D_x) + V_x.
}
\]

Clickbait texts are designed to maximize arousal while minimizing informational closure, creating a state of activated uncertainty that encourages user interaction which is captured by arousal. At the same time, it withholds key details or resolution, producing low dominance. Therefore, the term $A_h(1-D_h)$ models the core mechanism of curiosity as \textit{activated uncertainty}. When arousal is high and dominance is low, the reader experiences a stronger tendency to seek closure by clicking. Here, the term, $A_h(1-D_h)$, captures the interaction between emotional activation and lack of informational control, while the second term, $V_h$, reflects the contribution of positive affect. The curiosity amplification from the post to the stylized clickbait comment is then defined as
\[
\boxed{
\Delta CG^{(s)} =  CG(p) - CG(c^{(s)}) 
}
\]

or equivalently,

\[
\boxed{
\Delta CG^{(s)} =
\big[A_p(1-D_p) + V_p\big] - \big[A_{c^{(s)}}(1-D_{c^{(s)}}) + V_{c^{(s)}}\big]
}
\]

Here, $\Delta CG^{(s)}$ measures whether the stylized comment amplifies curiosity relative to the original post either positively or negatively.

Let $p$ denote the original post and $c^{(s)}$ denote a clickbait-stylized comment under style $s$. We define two key quantities: (i) emotional drift in VAD space, $\Delta E^{(s)}$, and (ii) curiosity amplification, $\Delta CG^{(s)}$.

A clickbait comment is considered effective when it can simultaneously increases curiosity while maintaining semantic and emotional coherence with the original post. Formally, this can be expressed as:

\[
\Delta CG^{(s)} \geq 0 \quad \text{or}
\quad \Delta CG^{(s)} < 0 
\]

The first condition ensures that the stylized comment amplifies the curiosity gap relative to the post, while the second condition constrains the emotional deviation to remain within a reasonable range. Therefore, the curiosity gap is maximized when a headline is emotionally activating, informationally unresolved, and framed accordingly. To make this precise, we define the effectiveness condition as positivie framing and negative framing:

\[
\boxed{
\Delta CG^{(s)} \geq 0 \quad postive \quad framing
}
\] and 

\[
\boxed{
\Delta CG^{(s)} < 0 \quad negative\quad framing
}
\]


A large positive $\Delta CG^{(s)}$ indicates that the comment introduces heightened arousal and reduced dominance (i.e., a stronger information gap), which are characteristic of clickbait. However, if $\Delta CG^{(s)}$ becomes too large, the comment may diverge significantly from the original post, reducing semantic coherence and potentially harming credibility. Therefore, an optimal clickbait strategy balances curiosity amplification is to maximize $\Delta CG^{(s)}$ or minimize $\Delta CG^{(s)}$ for emotional activation.


\begin{table*}[t]
\centering
\caption{Comparison of LLMs (e.g. Roberta and BERT) Performance Across Original Text and Stylized Rewrites of Clickbait Text }
\label{tab:model_comparison}

\resizebox{\textwidth}{!}{
\begin{tabular}{lcccc|cccc}
\hline
\multirow{2}{*}{\textbf{Text Type}} 
& \multicolumn{4}{c}{\textbf{Roberta }} 
& \multicolumn{4}{c}{\textbf{BERT}} \\
\cline{2-5} \cline{6-9}
& \textbf{Accuracy} & \textbf{Precision$^*$} & \textbf{Recall$^*$} & \textbf{F1} 
& \textbf{Accuracy} & \textbf{Precision$^*$} & \textbf{Recall$^*$} & \textbf{F1} \\
\hline
Original Text              & 0.9998 & 1.0000 & 0.9998 & 0.9999 & 0.9991 & 1.0000 & 0.9991 & 0.9995 
\\ \hline \hline 
Clickbait Rewrite     & 0.7370 & 1.0000 & 0.7370 & 0.8486 & 0.9770 & 1.0000 & 0.9770 & 0.9884 
\\ \hline
Neutral Rewrite       & 0.6362 & 1.0000 & 0.6362 & 0.7777 & 0.8927 & 1.0000 & 0.8927 & 0.9433 \\
Formal Rewrite        & 0.4683 & 1.0000 & 0.4683 & 0.6379 & 0.8527 & 1.0000 & 0.8527 & 0.9205 \\
Casual Rewrite        & 0.8027 & 1.0000 & 0.8027 & 0.8906 & 0.9864 & 1.0000 & 0.9864 & 0.9931 \\
Inspirational Rewrite & 0.5678 & 1.0000 & 0.5678 & 0.7244 & 0.9629 & 1.0000 & 0.9629 & 0.9811 \\
Humor Rewrite         & 0.7600 & 1.0000 & 0.7600 & 0.8636 & 0.9747 & 1.0000 & 0.9747 & 0.9872 \\
\hline
\end{tabular}
}

\vspace{2mm}
\footnotesize{Note: Precision is 1.0 and Recall is same as accuracy across all configurations because the evaluation dataset contains only positive clickbait samples, resulting in zero false positives. Therefore, accuracy, recall and F1-score provide meaningful insights into model performance under stylistic variations}
\end{table*}
\section{Approach}


\subsection{Dataset Selection} 
The primary dataset used in this study is the Clickbait Challenge 2017 Dataset. It is a widely adopted benchmark for clickbait detection tasks that consists of short textual content, primarily social media posts and headlines, collected from platforms such as Twitter. Each instance is paired with a human-annotated label indicating the degree of clickbait, typically represented in scores 0 (non-clickbait) and 1 (highly clickbait)~\cite{gollub2017clickbait}. For social media posts, we use the Pushshift Reddit Dataset, a large-scale publicly available corpus of Reddit posts and comments collected via the Pushshift API. The dataset contains millions of user-generated entries spanning diverse topics, which are available with communities, posts, and threaded comments in text format~\cite{baumgartner2020pushshift}. 

To integrate this dataset with our clickbait framework, we align Reddit posts with externally generated clickbait-style headlines using semantic similarity techniques. Specifically, each clickbait instance is matched to the most semantically relevant Reddit post using sentence embedding models, ensuring contextual consistency. This pairing enables us to simulate realistic clickbait insertion scenarios within organic discussions.

Furthermore, we annotate the resulting text pairs with emotion labels and map them into the Valence–Arousal–Dominance (VAD) space. Using these representations, we compute the curiosity gap (CG) and its variation ($\Delta CG$) across different stylistic transformations. This enriched dataset allows us to analyze how emotional and stylistic factors influence clickbait effectiveness and model robustness in a real-world social media setting.

\subsection{Dataset Merging} 
To construct a realistic evaluation setting, we combine the Clickbait Challenge 2017 Dataset with the Pushshift Reddit Dataset using a semantic similarity matching strategy. Specifically, we employ the Sentence-BERT (SBERT) model, implemented via the sentence transformers library, with the pre-trained model ``BAAI/bge-large-en-v1.5'' to encode both clickbait headlines and Reddit posts into dense vector representations.
Given a set of clickbait headlines $\{h_i\}$ and a pool of Reddit posts $\{p_j\}$, we compute cosine similarity between their embeddings. For each clickbait instance, we select the most semantically similar Reddit post as:

\begin{equation*}
p^{*} = \arg\max_{p_j} \ \cos\big(\text{SBERT}(h_i), \text{SBERT}(p_j)\big)
\end{equation*}

To ensure data diversity and avoid duplication, we enforce a one-to-one matching constraint, where each Reddit post is assigned to at most one clickbait headline. This results in a set of aligned pairs $(h_i, p^{*})$ that are semantically consistent. Due to the large size of the Pushshift dataset, we use the first 500,000 posts that have 118,447 valid title and post data which is merged it with valid clickbait text with above approach.

\subsection{Training the Classifier} 
We train two LLM models ``microsoft/deberta-v3-base''  and ``bert-base-uncased'' with the clickbait datasets containing both clickbait and nonclickbait. For DEBERTA and BERT model training, we use the full clickbait dataset, however, we utilize the only clickbait (15999 samples) discarding the other non-clickbait text for post matching. Thereafter, we combine the clickbait with the Reddit dataset with sBERT SentenceTransformer named ``BAAI/bge-large-en-v1.5'' based on the semantic similarity. The minimum and maximum cosine similarity scores, ranging from 0.5889 to 0.8923, indicate moderate to high semantic alignment between clickbait headlines and matched posts. 


\subsection{Stylization and Emotion Annotation} To generate stylistic variations of the aligned text pairs, we utilize the instruction-tuned language model microsoft's Phi-3~\cite{abdin2024phi}  where we use the medium-4k-instruct version which does not require fine tuning to do syphilitic conversion. Given an input text, the model produces multiple rewrites under different stylistic constraints, including neutral, formal, casual, inspirational, and humor-based tones. These rewrites preserve the core semantic content while altering the linguistic style, enabling controlled analysis of stylistic effects on clickbait behavior.

For emotion annotation, we employ the pre-trained model roberta-base-go\_emotions~\cite{lowe2022roberta}, which is fine-tuned on the GoEmotions dataset~\cite{demszky2020goemotions}. This model assigns emotion labels from a set of fine-grained categories to each text instance. The predicted emotion labels are subsequently mapped into the VAD space, allowing us to quantify the emotional characteristics of both original and stylized texts. These annotations are further used to compute the curiosity gap (CG) and its variation ($\Delta CG$), enabling a systematic analysis of how stylistic and emotional transformations influence clickbait effectiveness and model predictions.

\section{Results and Analysis}

We evaluate the effectiveness of emotion-aware clickbait generation by analyzing its impact on model performance under different stylistic transformations and different levels of curiosity gap ($\Delta CG$). 

\subsection{Impact of Stylization on Performance}

Table~\ref{tab:model_comparison} shows the performance of two LLMs, RoBERTa and BERT across the original text and multiple stylistic rewrites. Both models achieve good performance on the original dataset, with accuracy exceeding 99\%. However, performance decreases significantly once the same semantic content is rewritten in alternative styles.

For RoBERTa, accuracy decreases from 99.98\% on the original text to 73.7\% for clickbait rewrites, 63.6\% for neutral rewrites, and 46.8\% for formal rewrites. BERT shows a similar but less severe trend, where accuracy decreases from 99.91\% to 97.7\% for clickbait rewrites, 89.3\% for neutral rewrites, and 85.3\% for formal rewrites. Among the rewritten variants, casual and humor-based styles remain relatively more robust than other rewrites, although both still reduce performance compared to the original text. These findings suggest that stylistic transformation, can substantially affect classifier behavior. In particular, the results indicate that existing models rely not only on semantic content but also on surface-level stylistic patterns, making them vulnerable to emotionally motivated rewrites.

\begin{table}[t]
\centering
\caption{RoBERTa Performance Comparison Between Lowest and Highest $\Delta CG$ }
\label{tab:roberta_delta}
\scriptsize
\setlength{\tabcolsep}{2.5pt}
\renewcommand{\arraystretch}{1.05}

\resizebox{\columnwidth}{!}{
\begin{tabular}{lcccc}
\hline
\textbf{$\Delta CG$} & \textbf{Accuracy} & \textbf{Precision$^*$} & \textbf{Recall$^*$} & \textbf{F1} \\
\hline
Lowest  & 0.7272 & 1.0000 & 0.7272 & 0.8420 \\
Highest & 0.6937 & 1.0000 & 0.6937 & 0.8192 \\
\hline
\end{tabular}
}

\footnotesize{\textit{Note: Precision is 1.0 and Recall is same as accuracy due to the absence of negative samples in the evaluation dataset, resulting in zero false positives. Highest and lowest $\Delta CG$ refers to the positive and negative framing respectively.}}
\end{table}

\begin{table}[t]
\centering
\caption{BERT Performance Comparison Between Lowest and Highest $\Delta CG$}
\label{tab:bert_delta}
\scriptsize
\setlength{\tabcolsep}{2.5pt}
\renewcommand{\arraystretch}{1.05}

\resizebox{\columnwidth}{!}{
\begin{tabular}{lcccc}
\hline
\textbf{$\Delta CG$} & \textbf{Accuracy} & \textbf{Precision$^*$} & \textbf{Recall$^*$} & \textbf{F1} \\
\hline
Lowest  & 0.9742 & 1.0000 & 0.9742 & 0.9870 \\
Highest & 0.9474 & 1.0000 & 0.9474 & 0.9730 \\
\hline
\end{tabular}
}

\footnotesize{\textit{Note: Precision is 1.0 and Recall is same as accuracy due to the absence of negative samples in the evaluation dataset, resulting in zero false positives. Highest and lowest $\Delta CG$ refers to the positive and negative framing respectively.}}
\end{table}

\begin{figure}[t]
\centering

\includegraphics[width=0.9\columnwidth]{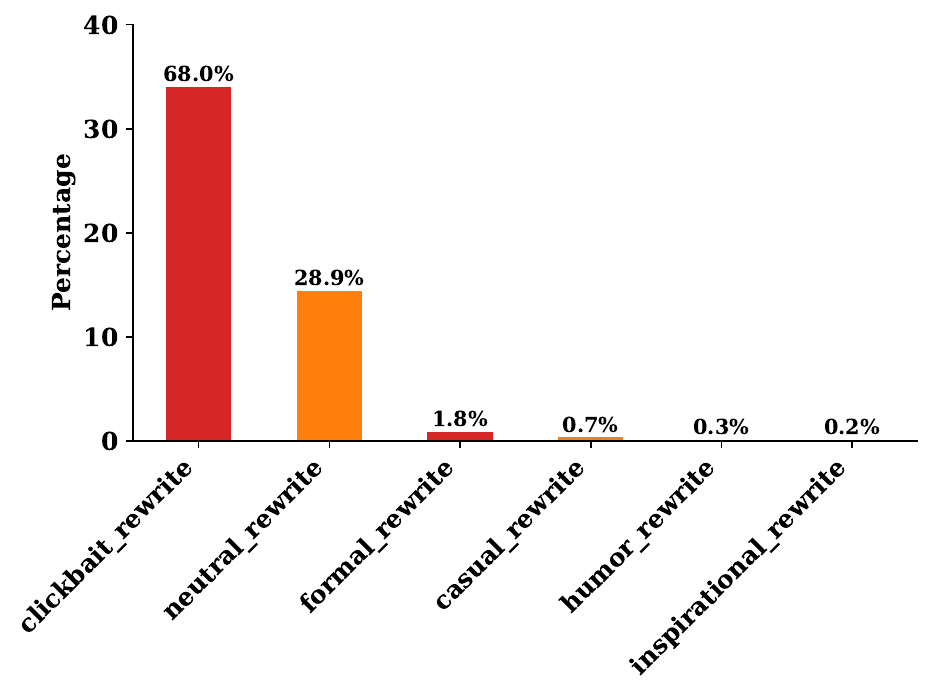}

{\small 1(a) Highest $\Delta CG$ }


\includegraphics[width=0.9\columnwidth]{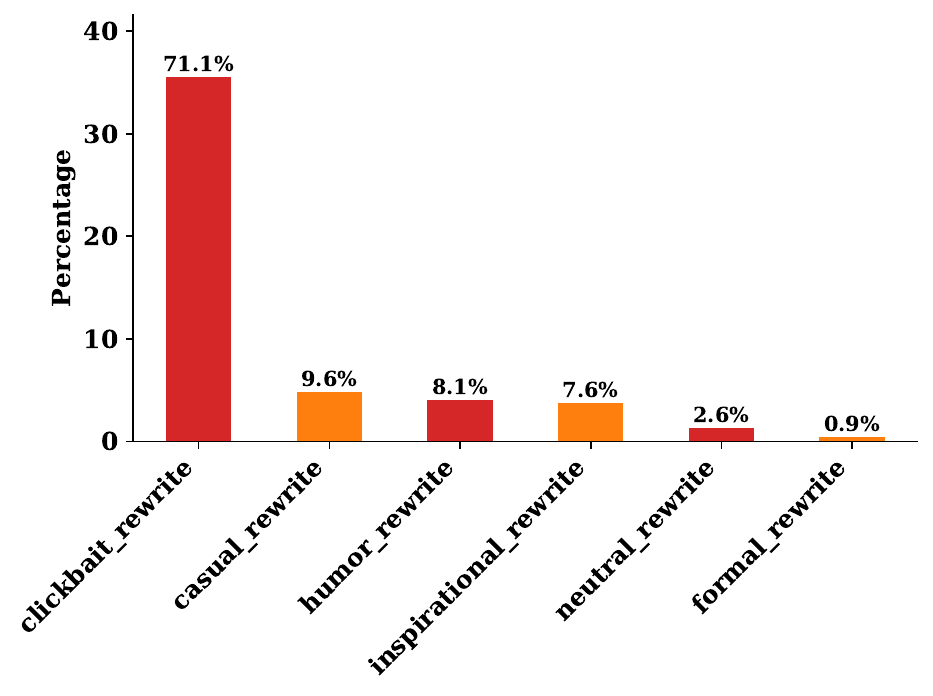}

{\small 1(b) Lowest $\Delta CG$}

\caption{Style distribution for highest and lowest $\Delta CG$ cases.}
\label{fig:highest_lowest}
\end{figure}

\subsection{Curiosity Gap on Classification }

To examine whether emotional amplification contributes to adversarial attack, we compare model performance on texts with the highest and lowest $\Delta CG$ values. Tables~\ref{tab:roberta_delta} and~\ref{tab:bert_delta} summarize these results.

For RoBERTa, accuracy drops from 72.72\% for the lowest $\Delta CG$ cases to 69.37\% for the highest $\Delta CG$ cases. A similar trend appears for BERT, where accuracy decreases from 97.42\% to 94.74\%. Although the reduction is moderate, the decline is consistent across both models indicate that texts with stronger curiosity amplification are harder for LLM to classify correctly. The higher $\Delta CG$ values correspond to a stronger adversarial effect, suggesting that  emotionally amplified clickbait increases the difficulty of detection.

\subsection{Style Distribution \& Curiosity Gap}

Figure~\ref{fig:highest_lowest} shows the distribution of rewriting styles among the highest and lowest $\Delta CG$ cases. For the highest $\Delta CG$ group, clickbait rewrites account for 68.0\% of the samples, followed by neutral rewrites at 28.9\%, while all other styles contribute only marginally. For the lowest $\Delta CG$ group, clickbait rewrites remain dominant at 71.1\%, followed by casual (9.6\%), humor (8.1\%), and inspirational (7.6\%) rewrites.

The high $\Delta CG$ cases are much more concentrated around clickbait, whereas low $\Delta CG$ cases are distributed more broadly across other stylistic categories. This indicates that strong curiosity amplification is most strongly associated with explicitly clickbait-oriented phrasing, while lower curiosity amplification allows greater stylistic diversity. Furthermore, the results demonstrate that the proposed emotion-aware clickbaiting strategy can produce measurable degradation in detection performance. Across styles and models, the performance reduction is approximately 3\% to 27\%, depending on the classifier. This validates that emotion-aware rewriting functions as an effective adversarial mechanism. 

The main insight from the results is that stylistic rewriting alone is sufficient to degrade clickbait detection performance, even when the underlying meaning is preserved. Secondly, larger curiosity-gap shifts are associated with increased misclassification, showing that emotional amplification contributes directly to adversarial success. Finally, clickbait-style rewrites dominate high $\Delta CG$ scenarios, confirming that emotionally charged, information-incomplete phrasing is the most effective strategy for inducing curiosity and reducing model robustness. The results also indicate that emotionally optimized clickbait can meaningfully affect classification performance and expose vulnerabilities in existing clickbait detection models. Moreover, the attack remains semantically aligned with the source post through SBERT-based matching while modifying the emotional framing through stylization. This makes the generated clickbait both contextually relevant yet difficult to detect, which increases the realism of the attack setting in social media environments.

\section{Limitation}

We acknowledge the following limitation for this research. Firstly, there is no publicly available dataset that jointly captures clickbait and social media posts, which limits the ability to evaluate the framework on fully integrated real-world data. Additionally, the stylistic transformations generated using the Phi-3 LLM may introduce bias, as the model is used without fine-tuning. Furthermore, the semantic similarity alignment relies on moderate embedding scores, and higher similarity would further improve contextual relevance in adversarial injection scenarios. Finally, the proposed approach has not been evaluated under real-world human interaction settings or its effectiveness in the presence of active user feedback or platform moderation.

\section{Conclusion}

This work redefines clickbait as an emotion-driven social amplification strategy in online networks, where adversarial agents generate emotionally optimized and semantically stylistic clickbait to increase user engagement. An attacker devises clickbait as a dynamic, emotion-driven behavior using the VAD framework and defines a curiosity gap (CG) function to quantify how emotional activation through positive or negative framing drives user curiosity and interaction. The methodology combines two datasets, the Clickbait Challenge and Pushshift Reddit dataset, using semantic matching with SBERT and stylistic rewrites to analyze how different tones affect clickbait strength. Experimental results show that emotionally stylized content can significantly influence model predictions and user engagement, even causing misclassification in trained filters in social media.

\section{Ethical Considerations }
This study does not include any harmful content to demonstrate the efficacy of emotion in clickbaiting. Furthermore, this research is for educational purposes only.

\bibliographystyle{ACM-Reference-Format}
\bibliography{sample-base}

\appendix

\section{APPENDIX}
This appendix presents additional materials to support the study.

\subsection{Generative AI Usage}
Our paper uses ChatGPT, QuillBot, and Grammarly as grammar checking and sentence refinement tools. Additionally, for some icon generation, we take help from Gemini. However, the authors are solely responsible for the experiments, analyses, and conclusion.

\subsection{Language Models}


In this study, we use the following LLMs to generate and analyze the output. Firstly, we use BERT style the pretrained model ``BAAI/bge-large-en-v1.5'' to compute semantic similarity between clickbait headlines and social media posts (ref table~\ref{tab:bge_default_config}). This allows us to align each generated or existing clickbait instance with the most contextually relevant post in the embedding space using cosine similarity. Secondly, we employ the instruction-tuned model Phi-3-mini-4k-instruct to generate multiple stylistic rewrites of the aligned text, including clickbait, neutral, formal, casual, inspirational, and humor-based variations while preserving the original semantic meaning (ref table~\ref{tab:phi3_config}). Finally, we utilize the pretrained roberta-base-go\_emotions model to extract fine-grained emotional labels from both the original and stylized texts (ref table~\ref{tab:roberta_config}) and use the deberta (ref table~\ref{tab:deberta_config}) and BERT (ref table~\ref{tab:bert_config}) model to classify clickbait.

\begin{table}[h!]
    \centering
    \caption{Configuration of BAAI/bge-large-en-v1.5}
    
    \begin{tabular}{|p{3cm}|p{4cm}|}
        \hline
        \textbf{Configuration} & \textbf{Value} \\
        \hline
        Pretrained Model  & BAAI/bge-large-en-v1.5 \\
        \hline
        Model Type        &  Encoder \\
        \hline
        Architecture      & Encoder-only \\
        \hline
        Hidden Size       & 1024 \\
        \hline
        Embedding Dimension & 1024 \\
        \hline
        Number of Layers  & 24 \\
        \hline
        Attention Heads   & 16 \\
    
        Max Sequence Length & 512 tokens \\
        \hline
        Similarity Metric & Cosine Similarity \\
        \hline
    \end{tabular}
\label{tab:bge_default_config}
\end{table}

\begin{table}[h]
    \centering
        \caption{DeBERTa Configuration}
    \begin{tabular}{|p{3cm}|p{4cm}|}
        \hline
        \textbf{Configuration} & \textbf{Value} \\
        \hline
        Pretrained Model  & microsoft/deberta-v3-base \\
        \hline
        Learning Rate     & 2e-5 \\
        \hline
        Batch Size        & 16 \\
        \hline
        Max Sequence Length & 64 \\
        \hline
        Weight Decay      & 0.01 \\
        \hline
        Optimizer         & AdamW  \\
        \hline
    \end{tabular}

    \label{tab:deberta_config}
\end{table}

\begin{table}[hb]
    \centering
        \caption{BERT Configuration}
    \begin{tabular}{|p{3cm}|p{4cm}|}
        \hline
        \textbf{Configuration} & \textbf{Value} \\
        \hline
        Pretrained Model  & bert-base-uncased \\
        \hline
        Learning Rate     & 2e-5 \\
        \hline
        Batch Size        & 16 \\
        \hline
        Max Sequence Length & 64 \\
        \hline
        Weight Decay      & 0.01 \\
        \hline
        Optimizer         & AdamW  \\
       
        \hline
    \end{tabular}

    \label{tab:bert_config}
\end{table}

\begin{table}[b]
    \centering
     \caption{Phi-3 mini 4k instruct Configuration for Stylization}
    \begin{tabular}{|p{3cm}|p{4cm}|}
        \hline
        \textbf{Configuration} & \textbf{Value} \\
        \hline
        Pretrained Model  & microsoft/Phi-3-mini-4k-instruct \\
        \hline
        Model Type        & Decoder-only Transformer \\
        \hline
        Context Length    & 4,096 tokens \\
        \hline
        Generation Method & Greedy Decoding \\
        \hline
        Temperature       & 0.0 \\
        \hline
        Top-p             & 1.0 \\
        \hline
        Max New Tokens    & 400 \\
    
        \hline
        Precision         & bfloat16  \\
        \hline
    \end{tabular}
   
    \label{tab:phi3_config}
\end{table}

\begin{table}[b]
    \centering
    \caption{RoBERTa Configuration for Emotion Extraction Trained on Google's GoEmotions Dataset}
    \begin{tabular}{|p{3cm}|p{4cm}|}
        \hline
        \textbf{Configuration} & \textbf{Value} \\
        \hline
        Pretrained Model  & SamLowe/roberta-base-go\_emotions \\
        \hline
        Model Type        & Encoder  \\
        \hline
        Task              & Emotion Classification \\
        \hline
        Number of Classes & 28 (GoEmotions) \\
    
        \hline
    \end{tabular}

    \label{tab:roberta_config}
\end{table}

\end{document}